\documentclass[letterpaper]{article} 
\usepackage{aaai23}  
\usepackage{times}  
\usepackage{helvet}  
\usepackage{courier}  
\usepackage[hyphens]{url}  
\usepackage{graphicx} 
\usepackage{natbib}  
\usepackage{caption} 
\usepackage{algorithm}
\usepackage{algorithmic}
\usepackage{subcaption}

\usepackage{newfloat}
\usepackage{listings}
\DeclareCaptionStyle{ruled}{labelfont=normalfont,labelsep=colon,strut=off} 
\floatstyle{ruled}
\newfloat{listing}{tb}{lst}{}
\floatname{listing}{Listing}
\usepackage{multirow}
\title{Remote Task-oriented Grasp Area Teaching By Non-Experts through Interactive Segmentation and Few-Shot Learning}

\author {Furkan Kaynar\textsuperscript{\rm 1}\equalcontrib, Sudarshan Rajagopalan\textsuperscript{\rm 2}\equalcontrib, 
Shaobo Zhou\textsuperscript{\rm 1}, 
Eckehard Steinbach\textsuperscript{\rm 1}}

\affiliations{\textsuperscript{\rm 1}Technical University of Munich\\ School of Computation Information and Technology\\  Department of Computer Engineering\\
Chair of Media Technology \\
Munich Institute of Robotics and Machine Intelligence (MIRMI)\\Munich, Germany \\
\textsuperscript{\rm 2}Madras Institute of Technology, Anna University, India\\
}

\begin{document}
%
\maketitle
\begin{abstract}

A robot operating in unstructured environments must be able to discriminate between different grasping styles depending on the prospective manipulation task. Having a system that allows learning from remote non-expert demonstrations can very feasibly extend the cognitive skills of a robot for task-oriented grasping. We propose a novel two-step framework towards this aim. The first step involves grasp area estimation by segmentation. We receive grasp area demonstrations for a new task via interactive segmentation, and learn from these few demonstrations to estimate the required grasp area on an unseen scene for the given task. The second step is autonomous grasp estimation in the segmented region. To train the segmentation network for few-shot learning, we built a grasp area segmentation (GAS) dataset with 10089 images grouped into 1121 segmentation tasks. We benefit from an efficient meta learning algorithm for training for few-shot adaptation. Experimental evaluation showed that our method successfully detects the correct grasp area on the respective objects in unseen test scenes and effectively allows remote teaching of new grasp strategies by non-experts.

\end{abstract}

\section{Introduction}
\label{sec:intro}

\begin{figure*}[!h]
\captionsetup[subfigure]{justification=centering}
\centering
\begin{subfigure}{0.7\columnwidth}
\includegraphics[width=\columnwidth]{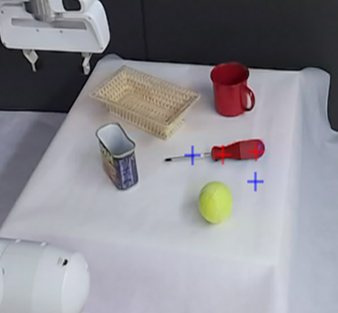}
\caption{Grasp area demonstrations via interactive segmentation. The selected region on the screwdriver handle is shown in red.}
\end{subfigure}\hspace{38 pt}
\begin{subfigure}{.7\columnwidth}
\includegraphics[width=\columnwidth]{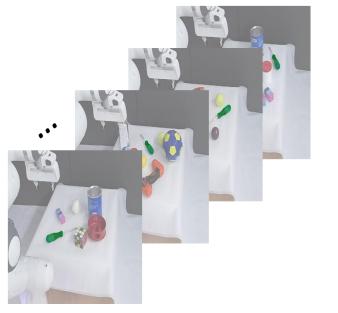}%
\caption{The obtained few-shot adaptation set after interactive segmentation. The ground truth masks are shown as green overlays.}
\end{subfigure}  \par\bigskip
\begin{subfigure}{.66\columnwidth}
\includegraphics[width=\columnwidth]{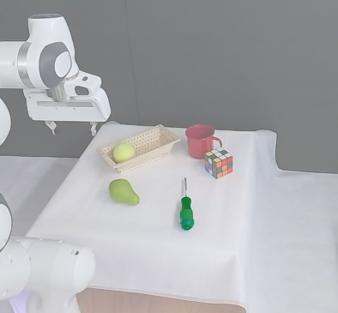}%
\caption{Predicted grasp area on a new scene after few-shot learning}
\end{subfigure}\hfill
\begin{subfigure}{.66\columnwidth}
\includegraphics[width=\columnwidth]{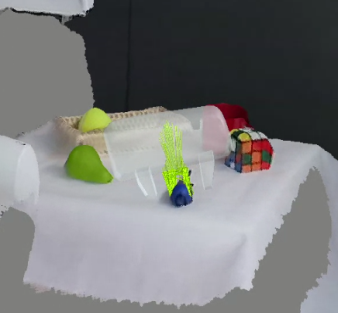}%
\caption{6-DOF grasp planning on the predicted grasp area}
\end{subfigure}\hfill
\begin{subfigure}{.66\columnwidth}
\includegraphics[width=\columnwidth]{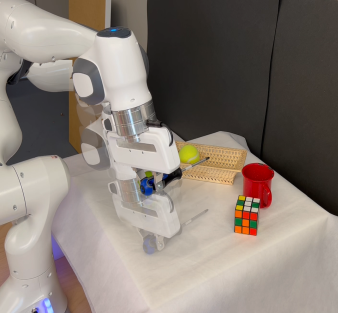}%
\caption{Grasp \\ execution}
\end{subfigure}
\caption{Pipeline of our remote grasp teaching framework. This illustration is for the task of grasping a screwdriver handle.}
\label{pipeline}
\end{figure*}

%
Robotic grasping of everyday objects in cluttered environments is a challenging problem. Many methods estimate grasps without considering the future tasks, but rather output grasps at an arbitrary part of the object that may not be suitable for the prospective manipulation task. To exemplify, for a pouring task, a household robot should not grasp a filled cup from the top, but rather from the side or the handle. The decision about the grasp location requires cognitive reasoning about the scene that we humans often do without even realizing.\let\thefootnote\relax\footnote{This work has been funded by the Lighthouse Initiative Geriatronics by StMWi Bayern (Project X, grant no. 5140951) and LongLeif GaPa gGmbH (Project Y, grant no. 5140953).}

There are data-driven task-oriented grasp estimation methods based on neural networks trained with large-scale datasets. There are two drawbacks of such approaches. First, creating large-scale datasets for task-oriented grasping is challenging due to the need for intensive human annotations. 
Second, the knowledge of the neural network cannot be updated easily for new grasping tasks as it often requires a complete new training with an even larger dataset. On the other hand, these issues can be avoided if the robot is able to learn new task-oriented grasp strategies from a few human demonstrations. Especially, being able to learn from non-expert guidance can greatly facilitate the deployment of robots in the unstructured environments like the household. 

The human guidance for task-oriented grasping can be provided in various ways. In this work, we focus on receiving human guidance remotely. The motivation for this is that the guiding person may not and does not have to physically access the robot or the objects for grasp demonstration. An example real-life use case is the elderly care robots operating in the household environment. These robots are supposed to help the seniors with the household tasks and improve their life quality. In many cases, the elderly person may not be the person responsible to guide the robot. In case of need for guidance for a new task, the robot needs to be operated remotely. If the robot is able to receive non-expert human guidance, any non-technical person can help the robot remotely, which makes the use of these robots more feasible in real life.

Existing few-shot grasp learning methods often require expert knowledge for kinesthetic demonstration \cite{kopicki2016one}, are limited to 3 or 4-DOF grasps (top grasps) \cite{van2018learning,yang2021attribute, guo2022few}, or they only work in a structured uncluttered environment with known or simple-shaped objects \cite{van2018learning}. Furthermore, the previous work often requires that the demonstrator is in the same physical place as the robot, or the demonstrator has to touch the objects to be able to demonstrate a grasp \cite{helenon2020learning, wang2021demograsp,saito2022task}. These drawbacks and requirements limit the practical applications of few-shot grasp teaching and do not allow remote guidance.

Another point to consider is the abilities of the non-expert persons providing grasp guidance. The previous work showed that remotely received grasp demonstrations lead to a higher success when the guidance demonstration is simple, has low degrees-of-freedom and is combined with autonomous grasp planning for the final grasp generation \cite{leeper2012strategies,kent2020leveraging}. On the other hand, directly executed 6-DOF grasp demonstrations tend to fail more often. 

All of the above mentioned points lead us to our proposed solution. We approach the remote few-shot grasp teaching problem with a novel framework considering non-expert abilities and 6-DOF grasp complexity. The proposed solution is based on two steps:

The first step is the specification of the allowed grasping area for a task on the RGB image. Towards this, we receive segmentation demonstrations from non-experts via an interactive segmentation interface. This brings two advantages. First, it does not require any expert knowledge about the robotic system or the technical details of the grasping task, but the non-expert person simply interacts with the RGB scene based on the life experience. Second, it allows to specify a feasible grasp area rather than discrete grasp locations. Having the entire grasp area improves kinematic feasibility of the task-oriented grasping task.

The segmentation interface receives positive clicks to select a region of an object and negative clicks to deselect them. Using a combination of the positive and negative clicks, a non-expert person can effortlessly segment the required grasp regions (see Figure~\ref{pipeline}(a)). Once the required region is selected, we save the segmentation mask as a sample for the given grasping task. The masks and their corresponding RGB images are then used as a support set for adapting our few-shot segmentation network to the given task.

The network is trained using a meta learning approach \cite{nichol2018reptile} to allow few-shot adaptation. For meta-training, we created a grasp area segmentation (GAS) dataset consisting of 10089 RGB images along with the ground truth segmentation masks, using a large-scale grasp dataset \cite{fang2020graspnet}.

After predicting the specific grasp area for the task, the second step of our solution is 6-DOF grasp pose estimation that is performed using a state-of-the-art grasp estimator \cite{sundermeyer2021contact} and the registered depth image. Thereby, we generate 6-DOF grasp poses that are limited only to the regions suitable for the prospective task. 

Experimental evaluations showed that our segmentation network is able to be trained with a few demonstrations on different cluttered scenes, and successfully estimates the task-related grasp area on unseen cluttered scenes. The tasks used for testing are usual household tasks such as grasping the handle or grip of a fork, a drill machine, a cup, a tennis racket or a screwdriver. Grasping tests with a robotic manipulator showed that our framework indeed facilitates task-oriented grasp learning from few non-expert demonstrations. With our pipeline, we aim to extend the cognitive skills of a robot via remote non-expert support.

Our contributions are summarized as follows:
\begin{itemize}
\item We incorporate state-of-the-art interactive segmentation to receive the grasp area demonstrations intuitively from non-experts. 

\item We leverage few-shot learning of part-based grasp area segmentations given for a prospective task, to be able to operate effectively after getting a few demonstrations.

\item For training the network using meta learning, we create a  grasp area segmentation (GAS) dataset with 10089 RGB images of cluttered scenes, grouped into 1121 grasp-area based segmentation tasks. 
\end{itemize}

\section{Related Work}
\label{sec:related}

\subsection{Autonomous Grasp Estimation without Learning from Demonstration}
Robotic grasping research has been focusing on autonomous grasp estimation since decades. Although the literature in the past focused more on object model-based and analytically computed grasps, the state-of-the-art methods are mostly deep learning-based. Recent techniques estimate 6-DOF grasps on point cloud data  \cite{sundermeyer2021contact,mousavian20196,fang2020graspnet,liang2019pointnetgpd}.
These methods aim to output different stable grasps on the raw scene, without considering the respective task. Unless limited to some region, they occasionally give invalid estimations like grasps at the sides of a table, or the visible part of the robotic arm itself in the scene. 

There are task-oriented autonomous grasp estimation methods, ranging from hand-crafted features and object affordances to the model-free deep learning-based techniques. Semantics based methods focus on estimating object affordances \cite{do2018affordancenet} for further grasp planning.
Liu et al.  proposed a context-aware grasp estimation method which receives a large variety of semantic features and estimates a task-oriented grasp \cite{liu2020cage}. 

The above mentioned methods are based on batch training of neural networks with thousands or millions of examples. Hence, they require a large dataset and are not designed for learning from few demonstrations.

\subsection{Grasp Learning from Few Demonstrations}

Few-shot learning of grasp estimations has attracted the attention of many robotic researchers as it allows teaching of new grasp strategies easily. Van Molle et al. proposed a CNN-based approach for one-shot grasp learning \cite{van2018learning}. Their method only works for top grasps on the same workspace, and is tested on simple block shaped objects. Hélénon et al. proposed a CNN-based method for few-shot learning of prohibited and authorised grasping locations in the industrial context \cite{helenon2020learning}. Their method requires a demonstration by a human who grasps the object with the thumb and index fingers covered with coloured pads.
Wang et al. proposed a method that estimates the robotic grasp pose by observing a few human hand-object interactions \cite{wang2021demograsp}. This method requires the shape completion of the known object and human hand pose estimation. The demonstrator has to interact with the object to be able to train the robot. Butler et al. proposed an interactive segmentation scheme for human-in-the-loop manipulation planning \cite{butler2017interactive}. Their method aims to support the planning during operation but does not include any learning for future operations. Yang et al. proposed a model that gets a text input describing the object and outputs a grasp estimation on the respective object \cite{yang2021attribute}. It only estimates top grasps and was mostly tested on simple objects like cuboid, cylinder, sphere. The closest work to ours is by Guo et al., which learns the grasping point of new objects with few examples, and is based on Model Agnostic Meta Learning (MAML)  \cite{guo2022few}. This method only outputs planar grasps and allows limited flexibility for robotic grasp planning as it outputs a single grasp point rather than a grasp area on the object. Saito et al. proposed grasp learning from observation, necessitating the demonstrator to interact with the objects \cite{saito2022task}. Kopicki et al. proposed a method for one-shot grasp learning on point clouds  \cite{kopicki2016one}. This method is based on kinesthetic teaching, which may not be appropriate for non-experts and it requires the demonstrator to be in the same physical location with the robot. 

In our framework, we aim to overcome the drawbacks of the above mentioned methods. Our method is suitable for non-experts since the demonstrations do not require any background on the robotic system, but rather they are annotations on a 2D color scene. It allows remote support since the demonstrator does not need to interact with the robot or the objects physically. We deploy 6-DOF grasp estimation, and are thereby able to operate in cluttered environments. Our network can successfully learn from a few examples to segment the correct grasping part of everyday objects.

\section{Methodology}
\label{sec:methodology}

Our pipeline is shown in Figure~\ref{pipeline} and it starts with getting the RGB image from the robot's camera and loading it into an interactive segmentation interface. A human demonstrator views the scene and segments a specific part of an object for the given task. An example could be segmenting the handle of a cup for pouring tasks. To process the RGB image and user annotations during demonstration, we use a custom web-based interface combined with a click-based interactive segmentation method \cite{sofiiuk2020f}, which facilitates quick refinement of the segmentation mask by adding further clicks on the object or the background. We get a few grasp area demonstrations for each task and save the masks for few-shot adaptation. After the few-shot adaptation, the network receives an unseen image including the same object (potentially at a different pose, partly occluded and in a different environment), and outputs a segmentation mask on the relevant region of this object for the task. 

For RGB image segmentation, we use the vanilla U-Net architecture \cite{ronneberger2015u} that we train using an efficient meta learning algorithm \cite{nichol2018reptile}. We create the GAS dataset for meta-training, meta-validation and meta-testing as explained in the next section.

Finally, we apply postprocessing on the output segmentation mask for outlier elimination.

\subsection{Grasp Area Segmentation Dataset Creation}
\label{dataset_section}

In order to train our network, we need a dataset consisting of RGB images of cluttered scenes, and task-oriented grasp area segmentation masks on the objects. For each segmentation task, we want different images with the same object and the respective segmentation masks showing the same grasp area on the target object for the given task. 

To create the GAS dataset, we benefit from a large scale grasping dataset named GraspNet-1Billion \cite{fang2020graspnet}. This dataset includes 3D reconstructed cluttered scenes that are produced using depth frames from different viewpoints. In addition, it includes the ground truth poses of all objects and nearly 1 billion non-colliding 6-DOF parallel gripper poses on the 3D scenes.

\begin{figure*}[!h]
\captionsetup[subfigure]{justification=centering}
\centering
\begin{subfigure}{.78\columnwidth}
\includegraphics[width=\columnwidth]{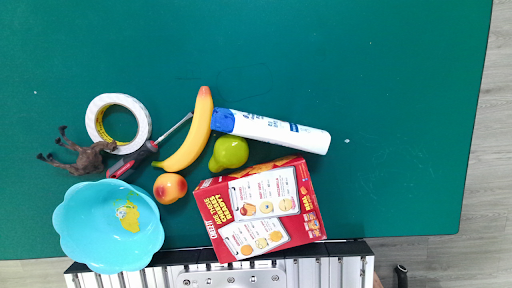}
\caption{RGB \\ scene}
\end{subfigure} \hspace{10 pt}
\begin{subfigure}{.78\columnwidth}
\includegraphics[width=\columnwidth]{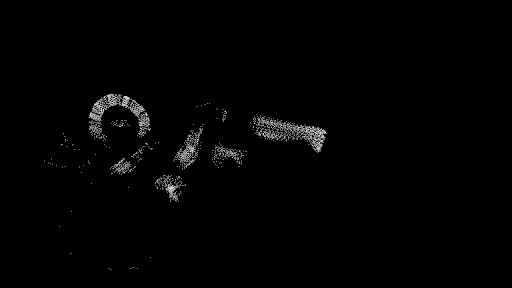}%
\caption{Grasp location map \\ for the given RGB scene}
\end{subfigure} \par\bigskip
\begin{subfigure}{.78\columnwidth}
\includegraphics[width=\columnwidth]{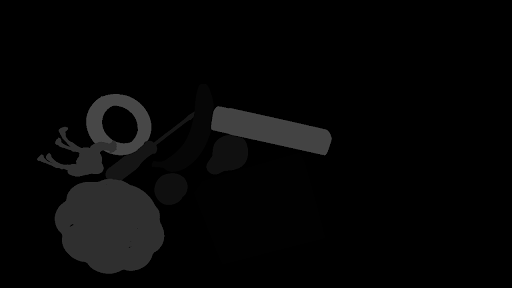}%
\caption{Ground truth\\ object masks}
\end{subfigure} \hspace{10 pt}
\begin{subfigure}{.78\columnwidth}
\includegraphics[width=\columnwidth]{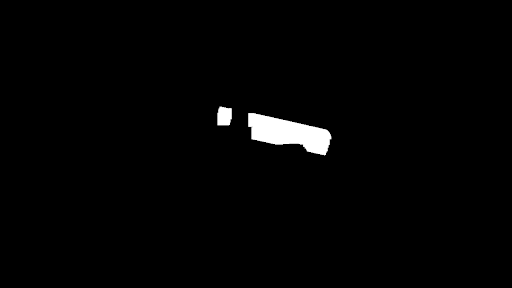}%
\caption{Grasp area segmentation mask\\ created for one object}
\end{subfigure}
\caption{An example of grasp area segmentation mask creation using data from GraspNet-1Billion dataset \cite{fang2020graspnet}.}
\label{dataset_creation_fig}
\end{figure*}

We create data for 1121 grasp segmentation tasks, each with 9 images, yielding 10089 RGB images and segmentation masks. An example of mask creation is shown in Figure~\ref{dataset_creation_fig}. The data is created using the following procedure:
\begin{itemize}

\item  For each reconstructed 3D scene in the GraspNet-1Billion dataset \cite{fang2020graspnet}, we get the source image sequence (256 images obtained by the Kinect Azure camera) and select images by skipping 20 frames in between. This gives us sufficiently different RGB images showing the same 3D scene arrangement. 
\item  For each RGB image of a scene, we get the provided extrinsic camera pose and the camera intrinsic matrix. 
\item On each 3D scene in the GraspNet-1Billion dataset \cite{fang2020graspnet}, the gripper poses are given in 6-DOF. We project the fingertip positions of each grasp onto the 2D RGB image pixel coordinates. Then, we average the two fingertip locations and save the center location on the RGB image as a grasp location. Thereby, we obtain a grasp location map on all feasible objects on each RGB image. An example grasp location map can be seen in Figure \ref{dataset_creation_fig}.b.

As we want grasp regions rather than individual grasp points, we combine the grasp points into grasp segmentation area masks. This is done as follows: 

\item We convolve each grasp location map with a 5x5 Gaussian kernel, then erode the resulting image with a 7x7 morphological structuring element and apply 2 dilations with a 15x15 structuring element. After thresholding the image, we get a binary mask with the feasible grasp regions on the object.

\item We multiply each binary mask with the related ground truth object segmentation mask (shown in Figure \ref{dataset_creation_fig}.c.) to separate the grasp areas for each object. This also limits the grasp area inside the object regions only (the area could have overshot the object boundaries after the dilation step).  At this point, we have a feasible grasp region mask for each object on each RGB image of each scene. 

\item We eliminate images with too small a grasp area, or with too diverse grasp locations that result in non-smooth grasp area masks. 
\end{itemize}

An example RGB image and the obtained grasp area segmentation mask can be seen in Figures \ref{dataset_creation_fig}.a and \ref{dataset_creation_fig}.d.

In the end, we obtain 1121 segmentation tasks, each task consisting of 9 RGB images and the 9 segmentation masks showing the grasp area for the corresponding object on the same scene. Note that we do not label the grasp regions semantically, but rather save the entire feasible grasp region on each object as a single sample. Since the grasps causing collision were already eliminated in GraspNet-1Billion dataset, the grasp areas in general correspond to only graspable parts of the objects rather than the entire object. Hence, our GAS dataset includes partial segmentations of each object showing feasible grasp regions. In total, we obtain 10089 RGB images and grasp area segmentation masks as our dataset.

\subsection{Training with Meta Learning}
\label{maml_section}

During training, we prepare the segmentation network for future few-shot adaptation based on Reptile meta learning algorithm \cite{nichol2018reptile}. This is an efficient implementation of the first-order MAML approach. In the original MAML algorithm \cite{finn2017model}, in each iteration, the network parameters are modified with few-shot adaptation based on gradient descent and then the network is trained using gradient descent on the overall loss. After training with this approach, the neural network has suitable initialization parameters such that it can be easily retrained with few examples for a new task, with a small number of adaptation steps, and generalizes well. However, this method requires a high computational cost during training, since the outer loop takes gradients of the inner loop gradients, leading to second order gradient computations. On the other hand, the Reptile algorithm has a single loop which applies multiple steps of the stochastic gradient descent for each task. In each iteration, the parameters of the network get closer to the optimal initialization for different tasks. Nichol et al. \cite{nichol2018reptile} show that the Reptile algorithm yields comparable performance to MAML  \cite{finn2017model}, at a smaller computational cost. Hence, we train our network with the Reptile algorithm using various loss functions, and report the performance results in the Experimental Evaluation section. 

\subsection{Outlier Elimination}
The initial tests indicated that our network successfully segments the relevant object part for grasping (true positive), occasionally accompanied by smaller isolated outlier mask regions in the binary image that are false positives. The largest region is almost always a true positive region. Therefore, we eliminate isolated regions having an area less than 50\% of the largest mask area. This step changes the segmentation performance marginally as reported in Table \ref{table:tasks} with the outl. elim. (outlier elimination) label. 

\begin{figure}[!h]
\centering
\begin{subfigure}{.46\columnwidth}
\includegraphics[width=\columnwidth]{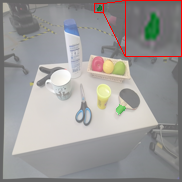}
\caption{Without outlier elimination}
\end{subfigure}\hfill
\begin{subfigure}{.46\columnwidth}
\includegraphics[width=\columnwidth]{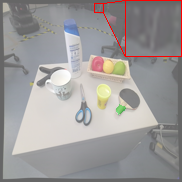}
\caption{With outlier elimination}
\end{subfigure}\hfill
\caption{Outlier elimination removes unwanted predictions as seen in this example. \textit{(a)}: The outlier prediction in green is indicated by a red box. The zoomed-in patch is provided in the top-right corner of the image; \textit{(b)}: The outlier is removed after applying our outlier elimination method.}
\label{fig:OE}
\end{figure}

Figure \ref{fig:OE} illustrates the effect of the outlier elimination on a prediction. The initially predicted mask includes a small false positive region in addition to the large true positive region on the tennis racket grip. Outlier elimination eliminates the false positive region and yields the true positive region.

\section{Experimental Evaluation}
\label{sec:experiment}

We use 1021 tasks in the GAS dataset for training, 50 tasks for validation and reserve 50 tasks for testing of our network. The training, validation and test sets include RGB images from different 3D scenes of the GraspNet-1Billion dataset \cite{fang2020graspnet}. In all the sets, each segmentation task includes 9 images. During training and validation, 5 images are randomly selected and used as few-shot training images and the remaining 4 are used for evaluating the mask estimation. The 5-shot training images are augmented with random rotation, random Gaussian blur, random color augmentation and random flipping during the meta-training. During validation and testing no augmentation was used on the few-shot training images.

\begin{figure}[!t]
\centering
\begin{subfigure}{.24\columnwidth}
\includegraphics[width=\columnwidth]{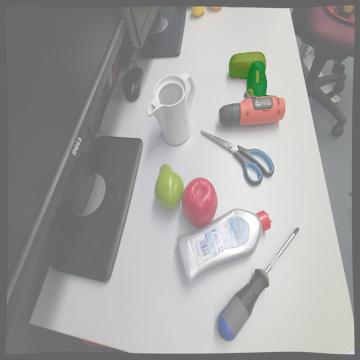}
\caption{}
\end{subfigure}\hfill
\begin{subfigure}{.24\columnwidth}
\includegraphics[width=\columnwidth]{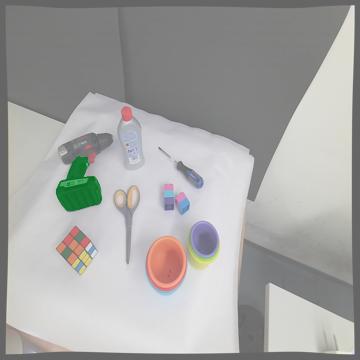}
\caption{}
\end{subfigure}\hfill 
\begin{subfigure}{.24\columnwidth}
\includegraphics[width=\columnwidth]{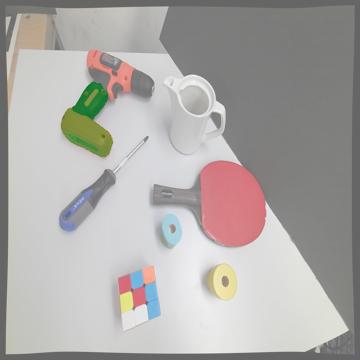}
\caption{}
\end{subfigure}\hfill
\begin{subfigure}{.24\columnwidth}
\includegraphics[width=\columnwidth]{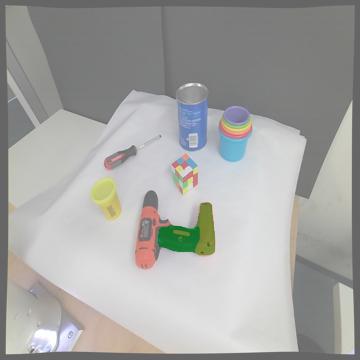}
\caption{}
\end{subfigure}\hfill  \\
\begin{subfigure}{.24\columnwidth}
\includegraphics[width=\columnwidth]{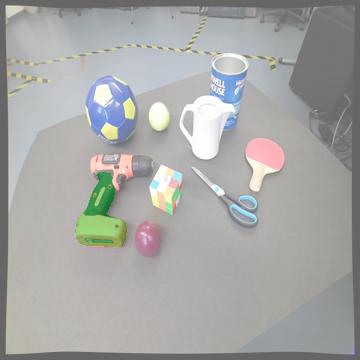}
\caption{}
\end{subfigure}\hfill 
\begin{subfigure}{.24\columnwidth}
\includegraphics[width=\columnwidth]{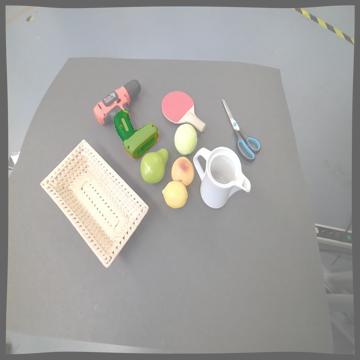}
\caption{}
\end{subfigure}\hfill
\begin{subfigure}{.24\columnwidth}
\includegraphics[width=\columnwidth]{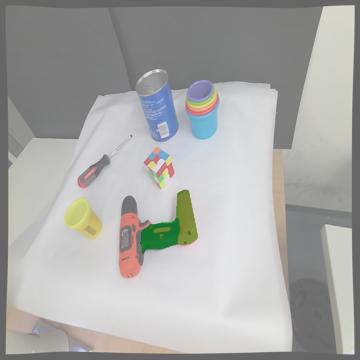}
\caption{}
\end{subfigure}\hfill
\begin{subfigure}{.24\columnwidth}
\includegraphics[width=\columnwidth]{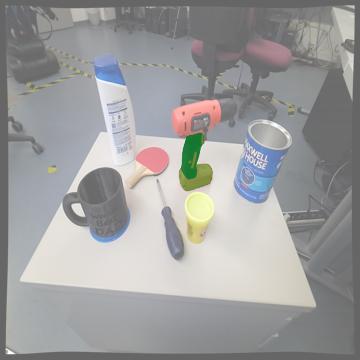}
\caption{} 
\end{subfigure}\hfill \\
\begin{subfigure}{.24\columnwidth}
\includegraphics[width=\columnwidth]{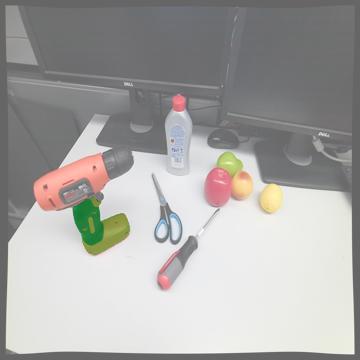}
\caption{}
\end{subfigure}
\begin{subfigure}{.24\columnwidth}
\includegraphics[width=\columnwidth]{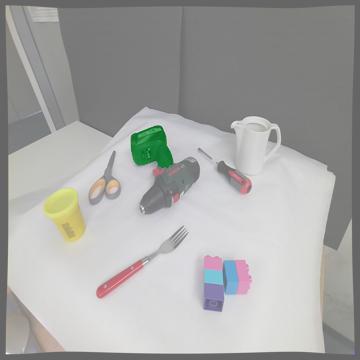}
\caption{}
\end{subfigure}\hfill
\caption{Example of a few-shot adaptation set with 10 shots for the task of grasping drill machine handle, from our custom test set.}
\label{fig:adaptation}
\end{figure}

\begin{figure*}[h!]
    \centering
    \setlength{\tabcolsep}{3pt}
    \begin{tabular}{c c c c c}
       (a) Cup Handle & (b) Drill Machine Grip & (c) Fork Grip & (d) Screwdriver Grip & (e) Tennis Racket Grip \\
       \includegraphics[width=0.395\columnwidth]{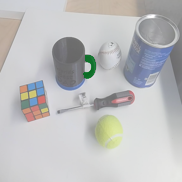} & \includegraphics[width=0.395\columnwidth]{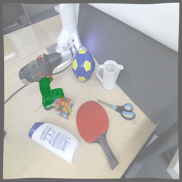} & \includegraphics[width=0.395\columnwidth]{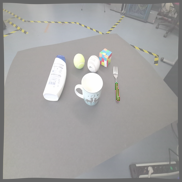} & \includegraphics[width=0.395\columnwidth]{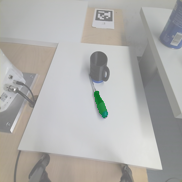} & \includegraphics[width=0.395\columnwidth]{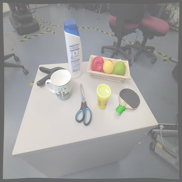} \\
       \includegraphics[width=0.395\columnwidth]{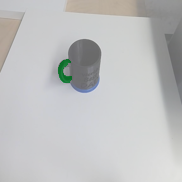} & \includegraphics[width=0.395\columnwidth]{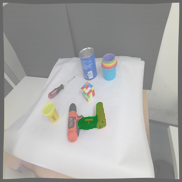} &
       \includegraphics[width=0.395\columnwidth]{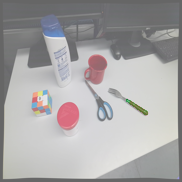} &  \includegraphics[width=0.395\columnwidth]{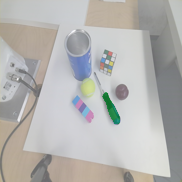}
        & \includegraphics[width=0.395\columnwidth]{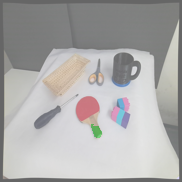}
    \end{tabular}
    \caption{Grasp-area prediction results using the network trained with the best configuration for various tasks from our custom test set. The predicted masks are shown as green overlay. Outlier elimination has been used for refining the predictions.}
    \label{customset}
\end{figure*}

In addition, we create a custom test set by taking new images with our RGB-D camera, leading to completely different scenes from the GAS dataset. An example few-shot adaptation set can be seen in Figure \ref{fig:adaptation}. With this custom set, we aim to test the generalization ability of our network.
The custom test set consists of 5 tasks which are common in a home-setting:

\begin{itemize}
    \item Cup Handle: Segment the handle
    \item Drill machine Grip: Segment the grip
    \item Fork Grip: Segment the grip
    \item Screwdriver Grip: Segment the grip
    \item Tennis racket Grip: Segment the grip
\end{itemize}

All images include cluttered scenes, and the target object is accompanied by other objects randomly. The camera pose is not fixed, hence the scenes can show different viewpoints of the objects. Example scenes of different tasks from our custom set can be seen in Figure \ref{customset}.

We use the mean Intersection-over-Union (mIOU) metric to evaluate the results on ground truth segmentation masks of the test sets. For each task of the custom test set, we use 10 images for few-shot adaptation and test the performance on 2 test images. Each task in the GAS test set includes 9 images. Hence, we use 8 images for few-shot adaptation and 1 image for testing.

\textbf{Implementation details.} We use a meta batch size and an inner loop batch size of 5. For the inner loop, we use the Adam optimizer, and for the outer loop we use stochastic gradient descent (SGD). We use a fixed learning rate of $3 \times 10^{-4}$ for the Adam optimizer and a meta learning rate decreasing from 1 to 0.001 for the SGD. Additionally we use a weight decay of $10^{-7}$ for both optimizers and set $\beta_{1}=0$, $\beta_{2}=0.999$ for the Adam optimizer. The number of adaptation steps during training, validation and testing are 5, 12 and 60, respectively.

\textbf{Analysis and discussions.} Training with binary cross entropy (BCE) loss with and without data augmentation showed that data augmentation improves the segmentation performance significantly on the custom test set (newly taken images), and marginally on the test set of GAS (including images from GraspNet-1Billion dataset). The difference between the performance improvements on the two test sets is expected, since the GAS' training set includes unseen scenes from GraspNet-1Billion, but with a similar scene arrangement to its test set.

It has been shown that, when the segmented area constitutes a small region in the masks, the learning can get stuck in a local minimum of the BCE loss function and the predictions can be significantly biased towards the background, resulting in a poor estimation of the foreground \cite{milletari2016v}. Milletari et al. proposed to use Dice's coefficient that mitigates this problem and improves the segmentation performance \cite{milletari2016v}. Following this principle, we train our network with a sum of Dice and BCE loss and observe that the performance improves significantly on the custom test set. 

Finally, the segmentation performance improves slightly after outlier elimination. The small difference indicates that our network already predicts a single connected component in the binary map that is on the correct part of the object.

\begin{table}[h!]
\centering
\caption{Average segmentation accuracy of the network using different training strategies.}
\begin{tabular}{|p{3.5cm}|p{1.6cm}|p{2.0cm}|}
 \hline
 \textbf{Method}& \textbf{mIoU}\newline \textbf{8-shot} \newline \textbf{(GAS set)} & \textbf{mIoU} \newline \textbf{10-shot} \newline \textbf{(Custom set)}  \\
 \hline
 BCE+No Aug.& 0.737 & 0.368  \\ 
 \hline
 BCE+Aug.& 0.743 & 0.489  \\ 
 \hline
 BCE+Dice+Aug. & 0.765 & 0.772  \\ 
 \hline
 BCE+Dice+Aug.+outl. elim. & 0.762& 0.806 \\
 \hline
\end{tabular}
\label{table:networks}
\end{table}

\begin{table}[h!]
\centering
\caption{10-shot segmentation accuracy on the individual tasks of the custom test set using the network trained with the best configuration.}
\begin{tabular}{|p{3.2cm}|p{1.2cm}|p{2.6cm}|}
 \hline
 \textbf{Tasks} & \textbf{mIoU}  & \textbf{mIoU} \newline \textbf{(with outl. elim.)} \\
 \hline
 Cup handle & 0.820 & 0.820 \\ 
 \hline
 Drill Machine Grip & 0.838 & 0.838 \\ 
 \hline
 Fork Grip & 0.799 & 0.844 \\ 
 \hline
 Screwdriver Grip  & 0.817 & 0.884\\
 \hline
 Tennis Racket Grip & 0.585 & 0.644 \\
 \hline
 \textit{Average (all tasks)} & 0.772 & 0.806\\
 \hline
\end{tabular}
\label{table:tasks}
\end{table}

\begin{table}[h!]
\centering
\caption{Comparison of segmentation accuracy for 1-shot, 5-shot and 10-shot adaptation on the custom set tasks, using the network trained with the best configuration.}
\begin{tabular}{|p{3.3cm}|c|c|c|}
 \hline
 \multirow{2}{*}{\textbf{Tasks}} & \multicolumn{3}{c|}{\textbf{mIoU}} \\ \cline{2-4}
 & \textbf{1-Shot}  & \textbf{5-Shot} & \textbf{10-Shot} \\
 \hline
 Cup Handle & 0.218 & 0.557 & 0.820 \\ 
 \hline
 Drill Machine Grip & 0.275 & 0.634 & 0.838 \\ 
 \hline
 Fork Grip & 0.122 & 0.886 & 0.844 \\ 
 \hline
 Screwdriver Grip & 0.411 & 0.866 & 0.884\\
 \hline
 Tennis Racket Grip & 0.127 & 0.55 & 0.644 \\
 \hline
 \textit{Average (all tasks)} & 0.230 & 0.698 & 0.806 \\
 \hline
\end{tabular}
\label{table:shots}
\end{table}

Table \ref{table:networks} shows the average prediction performance on the GAS test set and the custom test set,  for 8-shot and 10-shot learning, respectively. We use different training loss functions and analyze the results. The network trained with data augmentation and using the sum of the BCE and Dice losses performed the best. The results for the individual tasks of the custom test set are given in Table \ref{table:tasks}, for the network trained with BCE$ + $Dice loss and data augmentation. The qualitative results for the custom test set are shown in Figure~\ref{customset}.

Overall, we observe very good performance of grasp area prediction after adaptation using few samples. We also observed that the higher the variety in the few-shot adaptation set, the higher is the generalization performance of the networks. For instance, if the few-shot adaptation set includes only a single pose of an object, then the segmentation performance is lower on a test image that has the object at a different pose. A few-shot adaptation set with a lot of variety is shown in Figure \ref{fig:adaptation}.

We further analyze the performance of the network using different numbers of few-shot samples. Table \ref{table:shots} shows the outcome for the tasks in the custom test set. Even though we did not train our network for 1-shot learning, the mIoU values for 1-shot prediction are positive, showing that the estimated masks are partially overlapping with the correct grasp area. We observe that after 5-shot learning, the network already gives a reasonable output, which improves further for 10-shot learning. After 10-shot adaptation, the predicted masks are generally very close to the ground truth masks.

Finally, we demonstrate our framework on a robotic setup with a 7-DOF Franka Emika Panda manipulator, as shown in Figure~\ref{pipeline}. Here, the task is to grasp the handle of the screwdriver. We use a state-of-the-art 6-DOF grasp estimator \cite{sundermeyer2021contact} that can output grasps within a given segmentation area. We use ROS MoveIt! framework \cite{coleman2014reducing} for collision-aware motion planning and execution. We observe a successful task-oriented operation based on grasp areas that are learned from few non-expert demonstrations, showing the effectiveness of our method.

\section{Conclusion}
\label{sec:conclusion}

In this paper, we proposed a task-oriented grasp area learning framework based on remote non-expert demonstrations. We use interactive segmentation to receive simple demonstrations from non-experts intuitively. We leveraged a meta learning strategy to prepare the segmentation network for few-shot adaptation and segment the task-oriented grasp regions. We combine the grasp area segmentation with a state-of-the-art grasp planner to plan and execute the 6-DOF grasp only on the allowed grasp area. For training our network, we created the GAS dataset with 10089 RGB images and grasp area segmentation masks, using a large-scale grasp dataset. Experimental evaluation showed the success of our method for task-oriented grasp area learning from few remote non-expert demonstrations. Future work can focus on improving the network architecture and exploring different meta learning algorithms for this task.

--------------------------------------------------------------------

\bibliography{aaai23.bib}

\begin{thebibliography}{22}
\providecommand{\natexlab}[1]{#1}

\bibitem[{Butler, Elliot, and Cakmak(2017)}]{butler2017interactive}
Butler, D.~J.; Elliot, S.; and Cakmak, M. 2017.
\newblock Interactive scene segmentation for efficient human-in-the-loop robot
  manipulation.
\newblock In \emph{2017 IEEE/RSJ International Conference on Intelligent Robots
  and Systems (IROS)}, 2572--2579. IEEE.

\bibitem[{Coleman et~al.(2014)Coleman, Sucan, Chitta, and
  Correll}]{coleman2014reducing}
Coleman, D.; Sucan, I.; Chitta, S.; and Correll, N. 2014.
\newblock Reducing the barrier to entry of complex robotic software: a moveit!
  case study.
\newblock \emph{arXiv preprint arXiv:1404.3785}.

\bibitem[{Do, Nguyen, and Reid(2018)}]{do2018affordancenet}
Do, T.-T.; Nguyen, A.; and Reid, I. 2018.
\newblock Affordancenet: An end-to-end deep learning approach for object
  affordance detection.
\newblock In \emph{2018 IEEE international conference on robotics and
  automation (ICRA)}, 5882--5889. IEEE.

\bibitem[{Fang et~al.(2020)Fang, Wang, Gou, and Lu}]{fang2020graspnet}
Fang, H.-S.; Wang, C.; Gou, M.; and Lu, C. 2020.
\newblock Graspnet-1billion: A large-scale benchmark for general object
  grasping.
\newblock In \emph{Proceedings of the IEEE/CVF conference on computer vision
  and pattern recognition}, 11444--11453.

\bibitem[{Finn, Abbeel, and Levine(2017)}]{finn2017model}
Finn, C.; Abbeel, P.; and Levine, S. 2017.
\newblock Model-agnostic meta-learning for fast adaptation of deep networks.
\newblock In \emph{International conference on machine learning}, 1126--1135.
  PMLR.

\bibitem[{Guo et~al.(2022)Guo, Li, Hu, and Gan}]{guo2022few}
Guo, W.; Li, W.; Hu, Z.; and Gan, Z. 2022.
\newblock Few-Shot Instance Grasping of Novel Objects in Clutter.
\newblock \emph{IEEE Robotics and Automation Letters}.

\bibitem[{H{\'e}l{\'e}non et~al.(2020)H{\'e}l{\'e}non, Bimont, Nyiri, Thiery,
  and Gibaru}]{helenon2020learning}
H{\'e}l{\'e}non, F.; Bimont, L.; Nyiri, E.; Thiery, S.; and Gibaru, O. 2020.
\newblock Learning prohibited and authorised grasping locations from a few
  demonstrations.
\newblock In \emph{2020 29th IEEE International Conference on Robot and Human
  Interactive Communication (RO-MAN)}, 1094--1100. IEEE.

\bibitem[{Kent, Saldanha, and Chernova(2020)}]{kent2020leveraging}
Kent, D.; Saldanha, C.; and Chernova, S. 2020.
\newblock Leveraging depth data in remote robot teleoperation interfaces for
  general object manipulation.
\newblock \emph{The International Journal of Robotics Research}, 39(1): 39--53.

\bibitem[{Kopicki et~al.(2016)Kopicki, Detry, Adjigble, Stolkin, Leonardis, and
  Wyatt}]{kopicki2016one}
Kopicki, M.; Detry, R.; Adjigble, M.; Stolkin, R.; Leonardis, A.; and Wyatt,
  J.~L. 2016.
\newblock One-shot learning and generation of dexterous grasps for novel
  objects.
\newblock \emph{The International Journal of Robotics Research}, 35(8):
  959--976.

\bibitem[{Leeper et~al.(2012)Leeper, Hsiao, Ciocarlie, Takayama, and
  Gossow}]{leeper2012strategies}
Leeper, A.~E.; Hsiao, K.; Ciocarlie, M.; Takayama, L.; and Gossow, D. 2012.
\newblock Strategies for human-in-the-loop robotic grasping.
\newblock In \emph{Proceedings of the seventh annual ACM/IEEE international
  conference on Human-Robot Interaction}, 1--8.

\bibitem[{Liang et~al.(2019)Liang, Ma, Li, G{\"o}rner, Tang, Fang, Sun, and
  Zhang}]{liang2019pointnetgpd}
Liang, H.; Ma, X.; Li, S.; G{\"o}rner, M.; Tang, S.; Fang, B.; Sun, F.; and
  Zhang, J. 2019.
\newblock Pointnetgpd: Detecting grasp configurations from point sets.
\newblock In \emph{2019 International Conference on Robotics and Automation
  (ICRA)}, 3629--3635. IEEE.

\bibitem[{Liu, Daruna, and Chernova(2020)}]{liu2020cage}
Liu, W.; Daruna, A.; and Chernova, S. 2020.
\newblock Cage: Context-aware grasping engine.
\newblock In \emph{2020 IEEE International Conference on Robotics and
  Automation (ICRA)}, 2550--2556. IEEE.

\bibitem[{Milletari, Navab, and Ahmadi(2016)}]{milletari2016v}
Milletari, F.; Navab, N.; and Ahmadi, S.-A. 2016.
\newblock V-net: Fully convolutional neural networks for volumetric medical
  image segmentation.
\newblock In \emph{2016 fourth international conference on 3D vision (3DV)},
  565--571. IEEE.

\bibitem[{Mousavian, Eppner, and Fox(2019)}]{mousavian20196}
Mousavian, A.; Eppner, C.; and Fox, D. 2019.
\newblock 6-dof graspnet: Variational grasp generation for object manipulation.
\newblock In \emph{Proceedings of the IEEE/CVF International Conference on
  Computer Vision}, 2901--2910.

\bibitem[{Nichol and Schulman(2018)}]{nichol2018reptile}
Nichol, A.; and Schulman, J. 2018.
\newblock Reptile: a scalable metalearning algorithm.
\newblock \emph{arXiv preprint arXiv:1803.02999}, 2(3): 4.

\bibitem[{Ronneberger, Fischer, and Brox(2015)}]{ronneberger2015u}
Ronneberger, O.; Fischer, P.; and Brox, T. 2015.
\newblock U-net: Convolutional networks for biomedical image segmentation.
\newblock In \emph{International Conference on Medical image computing and
  computer-assisted intervention}, 234--241. Springer.

\bibitem[{Saito et~al.(2022)Saito, Sasabuchi, Wake, Takamatsu, Koike, and
  Ikeuchi}]{saito2022task}
Saito, D.; Sasabuchi, K.; Wake, N.; Takamatsu, J.; Koike, H.; and Ikeuchi, K.
  2022.
\newblock Task-grasping from human demonstration.
\newblock \emph{arXiv preprint arXiv:2203.00733}.

\bibitem[{Sofiiuk et~al.(2020)Sofiiuk, Petrov, Barinova, and
  Konushin}]{sofiiuk2020f}
Sofiiuk, K.; Petrov, I.; Barinova, O.; and Konushin, A. 2020.
\newblock f-brs: Rethinking backpropagating refinement for interactive
  segmentation.
\newblock In \emph{Proceedings of the IEEE/CVF Conference on Computer Vision
  and Pattern Recognition}, 8623--8632.

\bibitem[{Sundermeyer et~al.(2021)Sundermeyer, Mousavian, Triebel, and
  Fox}]{sundermeyer2021contact}
Sundermeyer, M.; Mousavian, A.; Triebel, R.; and Fox, D. 2021.
\newblock Contact-graspnet: Efficient 6-dof grasp generation in cluttered
  scenes.
\newblock In \emph{2021 IEEE International Conference on Robotics and
  Automation (ICRA)}, 13438--13444. IEEE.

\bibitem[{Van~Molle et~al.(2018)Van~Molle, Verbelen, De~Coninck, De~Boom,
  Simoens, and Dhoedt}]{van2018learning}
Van~Molle, P.; Verbelen, T.; De~Coninck, E.; De~Boom, C.; Simoens, P.; and
  Dhoedt, B. 2018.
\newblock Learning to grasp from a single demonstration.
\newblock \emph{arXiv preprint arXiv:1806.03486}.

\bibitem[{Wang et~al.(2021)Wang, Manhardt, Minciullo, Garattoni, Meier, Navab,
  and Busam}]{wang2021demograsp}
Wang, P.; Manhardt, F.; Minciullo, L.; Garattoni, L.; Meier, S.; Navab, N.; and
  Busam, B. 2021.
\newblock DemoGrasp: Few-shot learning for robotic grasping with human
  demonstration.
\newblock In \emph{2021 IEEE/RSJ International Conference on Intelligent Robots
  and Systems (IROS)}, 5733--5740. IEEE.

\bibitem[{Yang et~al.(2021)Yang, Liu, Liang, Lou, and Choi}]{yang2021attribute}
Yang, Y.; Liu, Y.; Liang, H.; Lou, X.; and Choi, C. 2021.
\newblock Attribute-based Robotic Grasping with One-grasp Adaptation.
\newblock In \emph{2021 IEEE International Conference on Robotics and
  Automation (ICRA)}. IEEE.

\end{thebibliography}

\end{document}